\pdfoutput=1

\documentclass[11pt]{article}

\usepackage[]{ACL2023}

\usepackage{times}
\usepackage{latexsym}
\usepackage{hyperref}

\usepackage[T1]{fontenc}

\usepackage[utf8]{inputenc}

\usepackage{microtype}

\usepackage{inconsolata}
\usepackage{hyperref} 
\usepackage{graphicx}

\title{Neural Machine Translation for Malayalam Paraphrase Generation }

\author{Christeena Varghese \\
  Technical University of Applied Sciences\\Würzburg-Schweinfurt\\Würzburg, Germany\\ \And
  Sergey Koshelev\\
  Institute of Linguistics, RAS\\
  Moscow, Russia\\
  \texttt{s.koshelev@iling-ran.ru}\\ \AND
  Ivan P. Yamshchikov\\
  CAIRO, \\Technical University of Applied Sciences\\Würzburg-Schweinfurt\\Würzburg, Germany\\ 
  \texttt{ivan.yamshchikov@thws.de}}

\begin{document}
\maketitle
\begin{abstract}
This study explores four methods of generating paraphrases in Malayalam, utilizing resources available for English paraphrasing and pre-trained Neural Machine Translation (NMT) models. We evaluate the resulting paraphrases using both automated metrics, such as BLEU, METEOR, and cosine similarity, as well as human annotation. Our findings suggest that automated evaluation measures may not be fully appropriate for Malayalam, as they do not consistently align with human judgment. This discrepancy underscores the need for more nuanced paraphrase evaluation approaches especially for highly agglutinative languages.
\end{abstract}

\section{Introduction}
Paraphrase generation is the task of rephrasing a given text while retaining its original meaning.  Paraphrase generation has attracted considerable attention in natural language processing (NLP) and computational linguistics. Alternatively, paraphrasing can be defined as rewriting a sentence in a different form without losing its semantic information. Thus automated paraphrasing is an essential component of any successful NLP system. For example, paraphrasing is essential for an NLP system to pass the Turing test. 

There are several notable ideas to generate paraphrases that are relevant in the context of this paper. First, the idea to use machine translation-inspired solutions for paraphrasing dates back to 
 \citet{quirk2004monolingual} who developed monolingual machine translation for paraphrase generation. Second, the idea of context-aware statistical paraphrase, that, to our knowledge, was first introduced in \citet{zhao2009application}.

Recently, \citet{li2017paraphrase} presented an approach that integrated deep reinforcement learning to obtain automated paraphrases. \citet{gupta2018deep} showed how deep generative networks could be used for paraphrase generation. Whereas \citet{egonmwan2019transformer} used transformers for paraphrase generation. \citet{zhou2021paraphrase} provide a more detailed view of paraphrase generation. 

In the context of linguistic diversity, the focus on paraphrasing extends beyond widely spoken languages to also include regional languages with rich linguistic nuances. \citet{salloum2011dialectal} addressed the challenge of dialectal to standard Arabic paraphrasing to enhance Arabic-English statistical machine translation. Their work signifies a critical effort to improve translation accuracy and fluency across different Arabic linguistic variants. Additionally, \citet{mizukami2014building} made a substantial contribution by creating a free, general-domain paraphrase database for the Japanese language. Furthermore,\citet{gao2018improving} explores the enhancement of English-to-Chinese neural machine translation through paraphrase-based data augmentation. 

This work addresses paraphrase generation in Malayalam. Malayalam is a Dravidian language spoken predominantly in Kerala. It is also spoken in Mahe and Lakshadweep of India, altogether resulting in a population of about 34 million. Malayalam is known for its complicated grammatical structures, complex verb conjugations, and extensive vocabulary.

There are several research projects addressing paraphrases identification Malayalam language and recognizing sentence similarities, see \cite{mathew2013paraphrase} and \cite{gokul2017sentence}. Recently, \citep{k2023abstractive} provided a Malayalam model for machine translation, text summarization, and question-answering.

As an extension of the above-mentioned studies, this paper aims to address the complex area of Malayalam paraphrase generation. Our investigation focuses on developing a specialized dataset tailored to Malayalam paraphrases, leveraging insights from established paraphrase generation models. The main motivation behind this research is to address the lack of resources for non-English languages and to improve the capabilities of NLP systems in the context of languages with special linguistic features.


\section{Related Works}
The generation of paraphrases is a longstanding problem for natural language learning, generating paraphrases in Malayalam is a relatively new task.

A seminal work by \citet{dolan2005automatically} emphasizes the importance of developing effective models for paraphrase generation, considering the varying syntactic and semantic expressions across different languages. These challenges become more pronounced in highly agglutinative languages, where words can be formed by stringing together multiple morphemes, adding an additional layer of complexity to the generation process. Extending paraphrase generation to Malayalam, a language with a complex linguistic structure, demands special attention. 

Scientific papers on multilingual NLP, such as the work by \citet{huang2020recent}, emphasize the need for language-specific adaptations in paraphrase generation models. The authors discuss the impact of linguistic diversity on the performance of NLP models, underscoring the importance of addressing language-specific challenges.

\citet{6921980} propose four similarity measures to predict the similarity between two sentences in Malayalam. Those are cosine similarity, Jaccard similarity, overlap coefficient, and containment measure. 

A shared Task on Detecting Paraphrases in Indian Languages, namely, Hindi, Tamil, Malayalam, and Punjabi are proposed by \citet{anand2018shared}. It consisted of two subtasks: Subtask 1 is to determine whether a sentence pair is a paraphrase or not, and Subtask 2 is to determine whether a sentence pair is a semi-paraphrase or a paraphrase or not. Different members use different features such as stop words, lemmatization, POS tagging, synonyms, overlap, cosine similarity, Jaccard similarity, etc. Due to the complexity of sentences, the F1 score and accuracy of Task 1 are comparatively high compared to the accuracy of Task 2. They concluded that the agglutinative character of Malayalam and Tamil makes paraphrasing more challenging.

\section{Data}

This paper uses the GYAFC  dataset \citet{rao-tetreault-2018-dear} in English as a start for the paraphrasing pipeline. This dataset consists of informal and formal sentence pairs which are built using the Yahoo Answers L6 corpus. The sentences in this dataset are obtained from various domains including Entertainment, Music, Family, Relationships, etc. Around 1000 English sentence pairs are available in this dataset.

Though we explore the possibility of adopting English datasets for Malayalam paraphrasing, we also provide a sample of 800 Malayalam paraphrase pairs evaluated by crowd workers with overlap of five\footnote{\url{https://github.com/Korzhinho/par_malayalam}}. The details on datalabelling are provided in Section \ref{eval}.

\section{Methods}

We try to explore four approaches that could potentially leverage the knowledge that we have for English and transfer it into Malayalam paraphrase. The first approach simply uses Google Translate on a random sample of 200 GYAFC paraphrases. We evaluate all four approaches on random 200 GYAFC sentence pairs.

The first model combines the output of Google Translate with MultiIndic Paraphrase Generation, a pre-trained model for paraphrase generation \citet{Kumar2022IndicNLGSM}. A prior study by \citet{zhou2018neural} served as the foundation for MultiIndic Paraphrase Generation, which extracts paraphrases from a parallel corpus. The model is developed using the Samanantar corpus \citet{ramesh2022samanantar}, which contains parallel corpora between English and all 11 Indic languages. 200 pairs of English phrases from the GYAFC dataset are translated into Malayalam using Google Translate. These translated Malayalam sentences are then fed into the MultiIndic Paraphrase Generation to obtain desired Malayalam paraphrase pairs. An illustrative example pertaining to this model can be found in Figure 1.

\begin{figure}[htp]
    \centering
    \includegraphics[width=8cm]{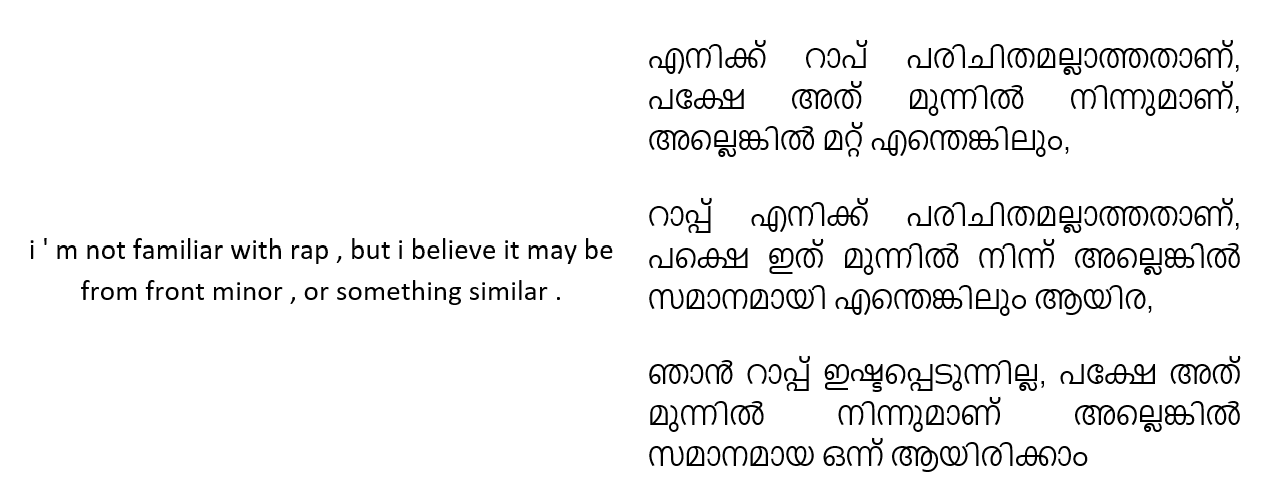}
    \caption{Result from Model 1}
    \label{fig: Result from Model 1}
\end{figure}

In the second approach, we use a set of English synonym word pairs\footnote{\url{https://github.com/i-samenko/Triplet-net/blob/master/data/data.csv}} to generate paraphrases in English with a simple synonym replacement heuristic approach to paraphrase. The generated paraphrases are then translated into Malayalam using Google Translate to obtain the Malayalam paraphrase set. Figure 2 exhibits an instance exemplifying this model, contributing to a deeper understanding.

\begin{figure}[htp]
    \centering
    \includegraphics[width=8cm]{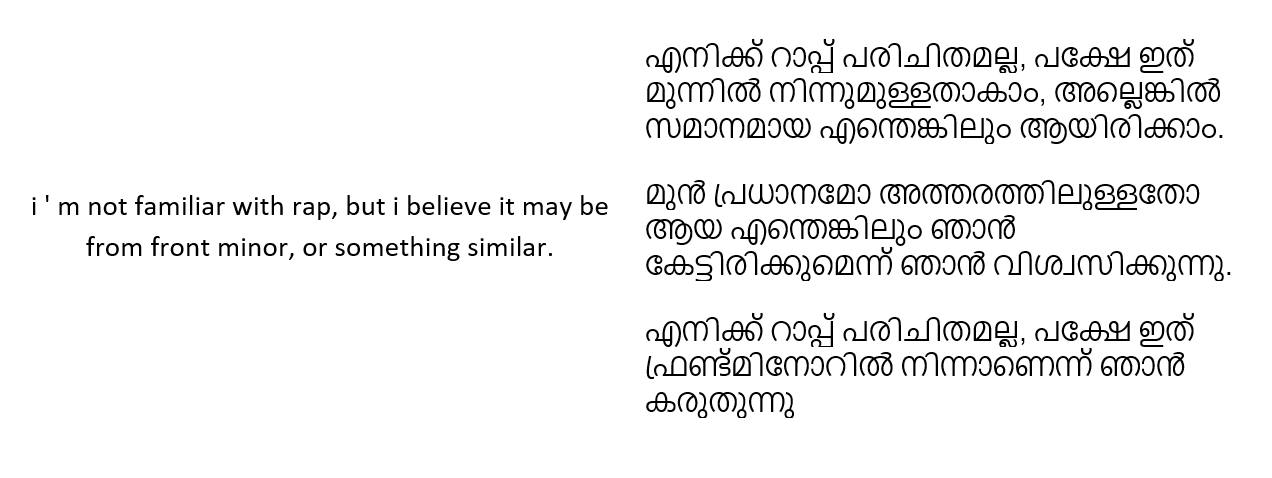}
    \caption{Result from Model 2}
    \label{fig: Result from Model 2}
\end{figure}

In the third approach, we use the bart-large-cnn model \citet{lewis2019bart}. Figure 3 contains an exemplar related to this model, offering additional clarity.

\begin{figure}[htp]
    \centering
    \includegraphics[width=8cm]{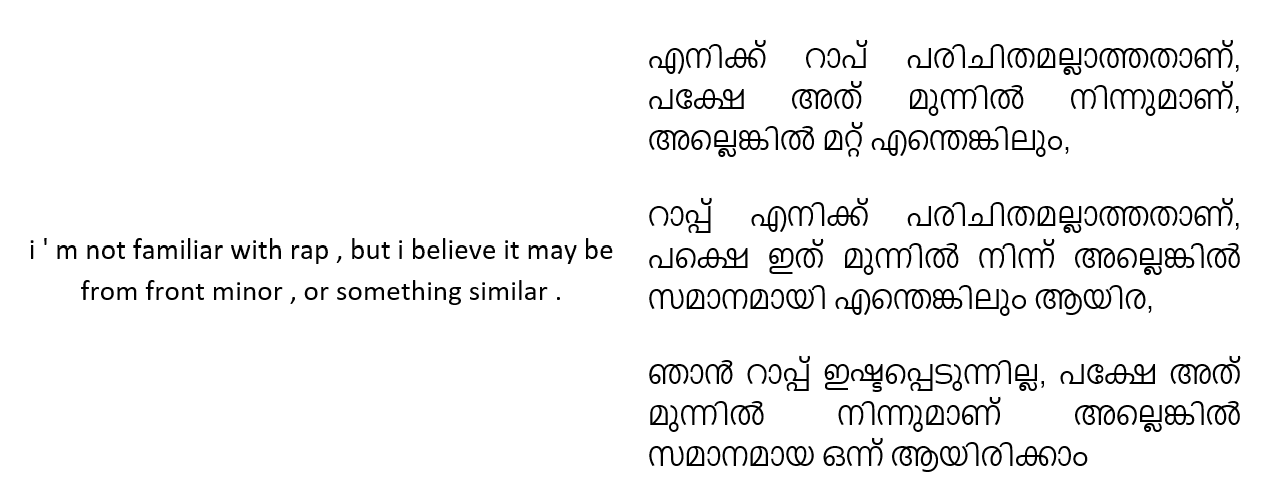}
    \caption{Result from Model 3}
    \label{fig: Result from Model 3}
\end{figure}

\begin{table*}[t!]
\centering
\begin{tabular}{lllll}
 \hline
 Model & BLEU & METEOR & cosine & human\\
& & & similarity & labels\\
 \hline 
 
 MultiIndic Paraphrase \cite{Kumar2022IndicNLGSM}& 0.04 & 0.25 & 0.70 & 0.37\\
 Synonym Replacement & 0.05 & 0.28 & 0.60 & \textbf{0.42} \\
 BART \cite{lewis2019bart} & 0.20 & 0.31 & \textbf{0.96} & 0.31\\ 
 OPUS \cite{tiedemann2012parallel} & \textbf{0.34} & \textbf{0.63} & 0.83 & 0.23\\
 Malayam Paraphrase \cite{anand2018shared} & \textbf{-} & \textbf{-} & 0.79 \footnotemark[3] & \textbf{-}
 
\end{tabular}

\caption{\label{tab:1}
Average BLEU score, METEOR score, Cosine Similarity as well as the percent of paraphrases labelled as correct paraphrase by human labellers for various models.
}
\end{table*}
\footnotetext[3]{The self-reported evaluation metric.}

Finally, in the fourth model a pre-existing language translation model named, OPUS(Open Parallel Corpus) \citet{tiedemann2012parallel}. OPUS models are a collection of pre-trained multilingual machine translation models developed by the Helsinki NLP group.  OPUS models are designed to handle translation tasks in several languages. They are trained to support translation between different language pairs, making them versatile for multilingual applications. Once again 200 pairs of sentences from the GYAFC dataset are passed to this model and Malayalam sentence pairs are generated. These translated sentences are then paraphrased by adjusting the beam-search parameters. Figure 4 includes an example associated with this model, providing supplementary clarity.

\begin{figure}[htp]
    \centering
    \includegraphics[width=8cm]{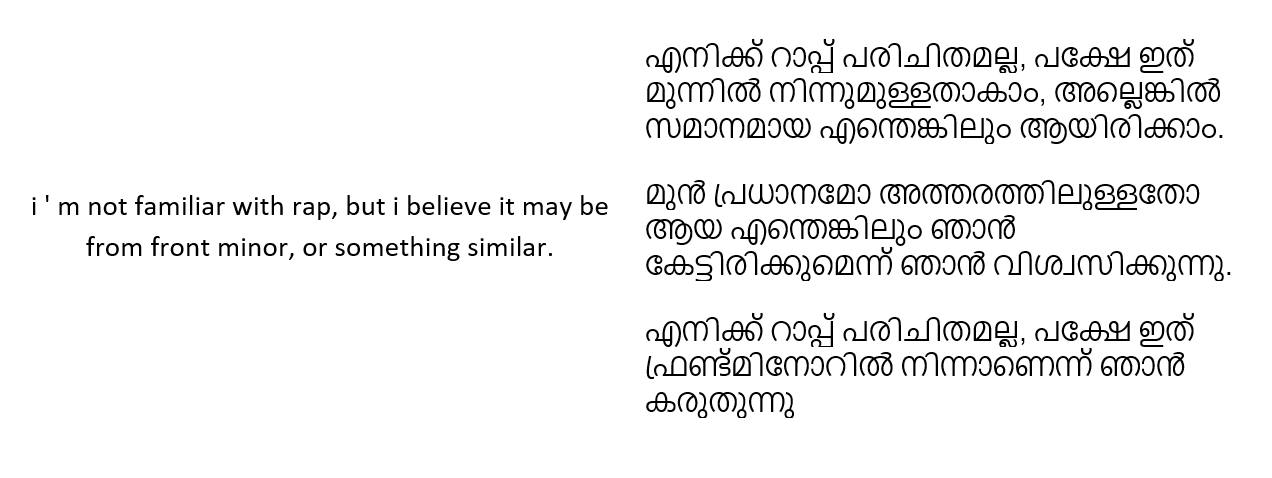}
    \caption{Result from Model 4}
    \label{fig: Result from Model 4}
\end{figure}

The num\_beams parameter controls the number of beams to use in beam search. Beam search is a decoding algorithm that explores multiple possible sequences and selects the most likely ones. A larger num\_beams value can increase diversity in generating phrases. Additionally, the num\_return\_sequences parameter determines how many different sequences to return. A higher value will result in more diverse paraphrases. Moreover, early\_stopping is used to speed up the paraphrase generation process. These parameters collectively influence the diversity, quality, and speed of paraphrase generation.

Finally, we compare these paraphrase methods based on NMT with the paraphrase proposed for Malayalam in \citet{anand2018shared}.

\section{Evaluation}\label{eval}


\citet{yamshchikov2021style} have explored various metrics for the evaluation of paraphrases. They found BERTScore \cite{zhang2019bertscore} to be the most adequate metric for English paraphrases. However, there is no direct analogy of BERTScore for Malayalam and the most commonly used metrics do correlate with human judgment \cite{solomon2022rethinking} on par with BERTScore though not perfectly.  Thus, in this work, we calculate the BLEU score \citep{papineni2002bleu} and METEOR score \citep{lavie2009meteor} for evaluating the phrases generated for a reference sentence. We also use cosine similarity used for paraphrase evaluation by \citet{anand2018shared} to put our results in perspective, despite cosine similarity was found to have a lower correlation with the human evaluation of paraphrases. Finally, we have labelled 200 paraphrase pairs generated by each of the models with human labellers via crowd-sourcing platform. Each sentence was labelled with three or more native speakers of Malayalam. We measured a percentage of sentence pairs that were labelled as correct paraphrases with high confidence. We publish the resulting human-labelled dataset of 800 sentence pairs to facilitate further research of paraphrasing in Malayalam.

Table \ref{tab:1} shows the results of the evaluation for 200 randomly sampled sentence pairs produced by four models that we test. It also puts these results into perspective comparing with the best result for Malayalam presented reported in \citet{anand2018shared} denoted in the Table \ref{tab:1} as 'Malayalam Paraphrase'.
 
 

One can see that the OPUS model outperforms other models in terms of automated evaluation metrics. In the meantime, the paraphrases generated with MultiIndic Paraphrase Generation, specifically designed for Indian languages, show lower results on automated evaluation. Comparison of the proposed methods with the best Malayalam paraphrasing model described in \citet{anand2018shared} also shows that on automated paraphrase evaluation metrics, direct application of machine translation methods, namely, BART or OPUS, leads to results that score higher in terms of BLEU, METEOR, and Cosine Similarity. However, this does not necessarily point at the weakness of the models but rather highlights the inadequacy of those popular evaluation metrics for Malayalam paraphrasing as well as the opportunity to leverage NMT to significantly expand the capabilities of Malayalam NLP. 

Once we include human evaluation into the picture we see two crucial results. First, the most successful paraphrases, according to human judgement, as simply achieved by heuristic synonym replacement. This is not surprising. What is important is that humans also evaluate MultiIndic Paraphrase higher that BART or OPUS, despite those models higher scores on automated metrics.

\section{Discussion}

In this study, we check if one could use machine translation methods for paraphrasing in Malayalam. We test several methods of generating paraphrases in English, followed by their translation into Malayalam. This methodology was compared with the performance of Malayalam-specific paraphrase models.

Our findings reveal that using English for initial paraphrase generation and then translating to Malayalam can yield results that are on par with those from Malayalam-specific models. This has several important implications:
\begin{itemize}
    \item Resource Optimization: This strategy showcases an efficient use of resources, leveraging the strengths of a high-resource language like English to benefit lower-resource languages;
    \item Model Versatility: The success of this approach suggests a potential shift in focus from developing language-specific models to enhancing translation-based methods;
    \item Expandability: such health check could be interesting for other Dravidian languages.
\end{itemize}

At the same time, one has to highlight certain limitations:
\begin{itemize}
    \item Translation Dependence: The effectiveness of paraphrases is heavily reliant on the accuracy and nuances captured by the machine translation process;
    \item Evaluation Metrics Concern: A critical limitation is the potential inadequacy of automated evaluation metrics in accurately capturing the quality of paraphrases in Malayalam. This raises concerns about the reliability of any paraphrase results solely evaluated automatically without any human labels whatsoever;
    \item Model Reliance: The approach's success is contingent on the performance of the English paraphrase models employed.
\end{itemize}

\section{Conclusion}

This study evaluates how effective is the idea to apply the existing neural machine translation methods to paraphrase generation in Malayalam. The core finding of this paper is that the models specifically designed for agglutinative languages like Malayalam are showing performance on par with NMT machine translation pipelines that leverage available English resources. The study also highlights the demand for specific paraphrase evaluation metrics more suitable for Dravidian languages. Finally, we publish human-labelled dataset of paraphrases to facilitate further research on the topic.

\bibliography{anthology,custom}
\bibliographystyle{acl_natbib}

\end{document}